\documentclass{article}

\PassOptionsToPackage{numbers, compress}{natbib}
\usepackage[preprint]{neurips_2025}

\usepackage[utf8]{inputenc}
\usepackage[T1]{fontenc}
\usepackage{hyperref}
\usepackage{url}
\usepackage{booktabs}
\usepackage{amsfonts}
\usepackage{nicefrac}
\usepackage{microtype}
\usepackage{xcolor}
\usepackage{amsmath}
\usepackage{graphicx}
\usepackage{wrapfig}
\usepackage{caption}
\usepackage{multirow}
\usepackage{enumitem}
\usepackage{subcaption}
\usepackage{svg}
\usepackage[most]{tcolorbox}

\title{Multimodal Forecasting of Sparse Intraoperative Hypotension Events Powered by Language Model}

\author{
\textbf{Jintao Zhang\textsuperscript{1}} \quad
\textbf{Zirui Liu\textsuperscript{1}} \quad
\textbf{Mingyue Cheng\textsuperscript{1}} \quad
\textbf{Shilong Zhang\textsuperscript{1}} \quad
\textbf{Tingyue Pan\textsuperscript{1}} \\
\textbf{Yitong Zhou\textsuperscript{1}} \quad
\textbf{Qi Liu\textsuperscript{1}}\thanks{Corresponding author.} \quad
\textbf{Yanhu Xie\textsuperscript{2}} \\
\textsuperscript{1}University of Science and Technology of China \\
\textsuperscript{2}The First Affiliated Hospital of University of Science and Technology of China \\
\{zjttt, liuzirui\}@mail.ustc.edu.cn, mycheng@ustc.edu.cn, zhangshilong@mail.ustc.edu.cn, \\ \{pty12345, yitong.zhou\}@mail.ustc.edu.cn, qiliuql@ustc.edu.cn, xyh200701@sina.cn
}

\begin{document}

\maketitle

\vspace{-0.25in}
\begin{abstract}
\vspace{-0.05in}


Intraoperative hypotension (IOH) frequently occurs under general anesthesia and is strongly linked to adverse outcomes such as myocardial injury and increased mortality. Despite its significance, IOH prediction is hindered by event sparsity and the challenge of integrating static and dynamic data in diverse patients. In this paper, we propose \textbf{IOHFuseLM}, a multimodal language model framework. To accurately identify and differentiate sparse hypotensive events, we use a two-stage training strategy. The first stage involves domain-adaptive pretraining on IOH physiological time series augmented through diffusion methods, thereby enhancing the model sensitivity to patterns associated with hypotension. Subsequently, task fine-tuning is performed on the original clinical dataset to further enhance the ability to distinguish normotensive from hypotensive states. To enable multimodal fusion for each patient, we align structured clinical descriptions with the corresponding physiological time series at the token level. This alignment enables the model to capture individualized temporal patterns alongside their corresponding clinical semantics. In addition, we convert static patient attributes into structured text to enrich personalized information. Experimental evaluations on two intraoperative datasets demonstrate that IOHFuseLM outperforms established baselines in accurately identifying IOH events, highlighting its applicability in clinical decision support scenarios. Our code is publicly available to promote reproducibility in \url{https://github.com/zjt-gpu/IOHFuseLM}.
  
\end{abstract}

\vspace{-0.15in}
\section{Introduction}
\vspace{-0.1in}

Intraoperative hypotension (IOH) is a common complication during surgery and has been consistently associated with adverse postoperative outcomes~\cite{I1,I2}, including myocardial injury~\cite{I3} and increased mortality~\cite{I4}. Given its high prevalence and substantial clinical implications~\cite{I5}, the development of accurate IOH prediction models has become a critical objective in perioperative monitoring~\cite{I6}.

Conventional approaches to IOH prediction are based primarily on physiological characteristics such as arterial blood pressure~\cite{R1}, and increasingly incorporate deep learning models to capture temporal dependencies in time series data~\cite{R4}. These methods typically leverage convolutional neural networks to extract local series patterns~\cite{I11}, or employ recurrent architectures based on attention to model sequential dynamics~\cite{I12}, achieving moderate performance gains. However, most existing methods focus exclusively on physiological time series~\cite{R5,I23} or adopt simple feature-level fusion strategies by concatenating static patient attributes~\cite{R6}, without fully modeling the semantic and contextual complexity of individual patients.

Recent advances in deep learning for time series forecasting have led to notable progress across diverse domains, with models such as LSTM~\cite{lstm}, Transformer-based~\cite{Crossformer, iTransformer, Timexer}, and MLP-based architectures~\cite{DLinear, FreTS} demonstrating strong capabilities in modeling temporal dynamics. Beyond deterministic models, diffusion-based generative approaches are effective for time series analysis~\cite{Diffusion-ts, RATD} and data augmentation via realistic sample synthesis~\cite{I31}. More recently, language models~\cite{GPT4TS, Timellm} have expanded the field of time series forecasting by enabling cross-modal representation learning and effectively aligning textual and temporal features.

\begin{figure}
    \centering
    \includegraphics[width=\textwidth]{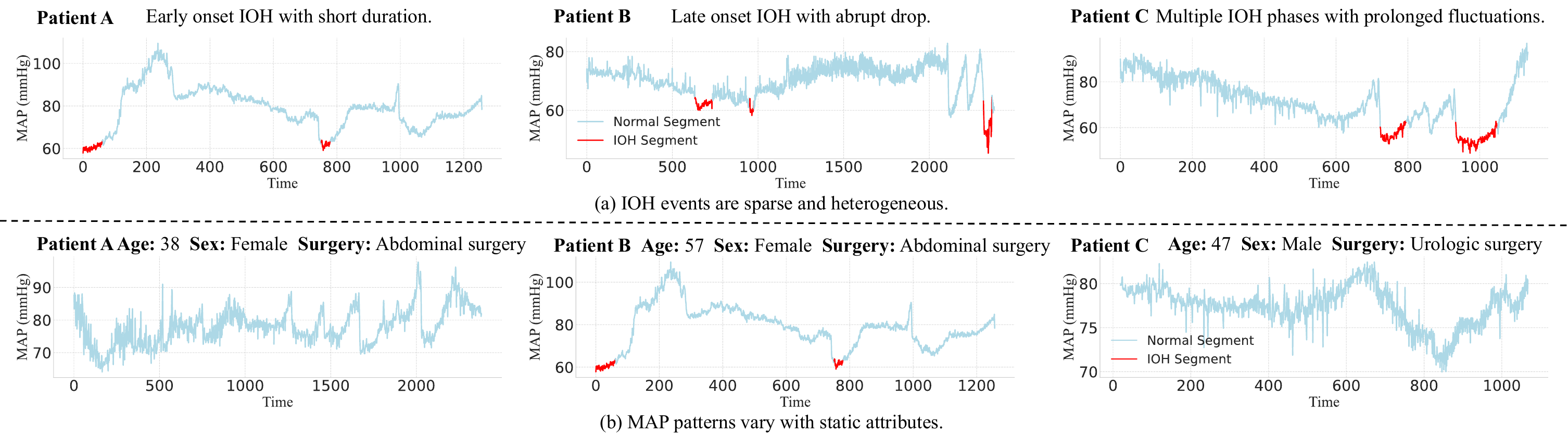}  
    \vspace{-0.15in}
    \caption{(a) IOH events are sparse and exhibit substantial inter patient variability in onset time, duration, and waveform morphology. (b) MAP series vary significantly across static attributes including age groups, genders and surgery types.}
    \vspace{-0.15in}
    \label{Motivation}
\end{figure}

Despite substantial progress, predicting IOH remains challenging~\cite{I21}. As shown in Figure~\ref{Motivation}~(a), IOH events are sparse, brief, and highly variable in onset time, waveform morphology, and temporal dynamics. Figure~\ref{Motivation}~(b) further shows that MAP fluctuations vary significantly between patients due to factors such as age and type of surgery, making it difficult for models to generalize across diverse populations. Thus, effective IOH modeling requires capturing critical temporal patterns while jointly integrating static patient attributes~\cite{I22} with dynamic physiological signals.

To address the challenge posed by sparse IOH events, we propose a multimodal language model framework, \textbf{IOHFuseFM}, which integrates static patient attributes with dynamic physiological series. The training process consists of two stages. First, domaim adaptive pretraining is conducted on a dataset augmented with a diffusion strategy to capture a diverse range of fine-grained patterns. This is followed by task fine-tuning using a customized loss function that improves the model’s sensitivity to IOH-related abnormalities. Static attributes are transformed into clinically informed descriptions, enabling cross-modal alignment through token-level interaction for precise semantic fusion.

Our main contributions are summarized as follows:
\vspace{-0.05in}
\begin{itemize}
    \item We propose IOHFuseLM, a novel multimodal language model framework for IOH prediction. The model is trained using a two-stage paradigm: domaim adaptive pretraining on a diffusion-augmented physiological dataset, followed by task fine-tuning on real intraoperative records.
    \item We develop a clinically informed multimodal fusion strategy that aligns static patient context with temporal physiological series by converting patient attributes into structured clinical text and aligning it at the token level with physiological series.
    \item Experiments on two real-world intraoperative datasets demonstrate that our method consistently outperforms competitive baselines, including a curated dataset based on raw intraoperative blood pressure recordings.
\end{itemize}

\vspace{-0.15in}
\section{Related Work}
\vspace{-0.05in}
\paragraph{Intraoperative Hypotension Forecasting.}  
Modeling intraoperative arterial pressure has emerged as a key strategy for early prediction of intraoperative hypotension (IOH), enabling timely clinical interventions and improved patient safety.
Early efforts primarily focused on high fidelity arterial pressure series, leading to the development of the Hypotension Prediction Index~\cite{R1}. Subsequent machine learning approaches, including ensemble methods~\cite{R2} and gradient boosting techniques~\cite{R3}, integrated both preoperative and intraoperative variables. However, these models often treated each data point in isolation, overlooking the intrinsic temporal dependencies. To address this limitation, deep learning architectures, including recurrent neural networks~\cite{R5} and attention-based models~\cite{R6}, were introduced to better capture sequential patterns. More recently, interpretable models~\cite{R7, R8} have improved clinical utility, although they still depend on predefined features and structured inputs. Meanwhile, the frequency-domain perspective~\cite{R9} has also been explored.
While existing IOH prediction methods have made considerable progress, most are grounded in either biomarker identification or deep learning models that lack the capacity to align patient specific clinical narratives with the evolving temporal dynamics of physiological series. Despite progress, existing IOH prediction methods still lack the capacity to bridge multimodal disparities and effectively model personalized, temporally evolving risk patterns.

\vspace{-0.05in}
\paragraph{Time Series Forecasting.}  
Time series forecasting~\cite{cheng2025comprehensive} plays a pivotal role in domains such as healthcare, weather, and energy. Classical statistical models, including ARIMA~\cite{Arima}, often struggle to capture the complex dynamics present in physiological series that exhibit high dimensionality and nonlinearity. Deep learning models, such as long short term memory networks~\cite{lstm}, convolutional neural networks~\cite{cheng2025convtimenet} and gated recurrent units~\cite{gru}, have demonstrated strong capabilities in modeling temporal dependencies over extended time horizons by leveraging gated mechanisms.
In recent years, Transformer-based architectures~\cite{Informer, Autoformer, Fedformer, cheng2023formertime} have achieved notable progress in time series forecasting. For instance, PatchTST~\cite{PatchTST} introduces patch-level embeddings, providing a principled approach to tokenizing time series. MLP-based models ~\cite{DLinear, Tsmixer, nips3, nips4} have also shown competitive performance with reduced computational complexity, while effectively preserving temporal structures through simplified model designs.

In addition to architectural advancements, generative modeling has emerged as a promising paradigm for time series forecasting, with diffusion-based approaches gaining increasing attention. Recent models~\cite{CSDI,Timediff, nips5,zhang2024fdf} effectively capture complex temporal dynamics by iteratively denoising noise-perturbed series through learned reverse processes.
At the same time, large language models have exhibited increasing potential in time series modeling~\cite{GPT4TS,Timellm,S2IP-llm, nips7}. Through pretraining and instruction tuning, LLMs are capable of generalizing forecasting capabilities across a wide range of tasks and domains, thereby enabling more flexible and adaptive series understanding. These advancements establish a solid foundation for developing unified and generalizable time series forecasting frameworks that combine high representational capacity with strong adaptability to the intricate dynamics characteristic of IOH prediction scenarios.

\vspace{-0.1in}
\section{Preliminaries}
\vspace{-0.1in}

\begin{wrapfigure}{r}{0.48\textwidth}
    \vspace{-0.1in}  
    \centering
    \includegraphics[width=0.48\textwidth]{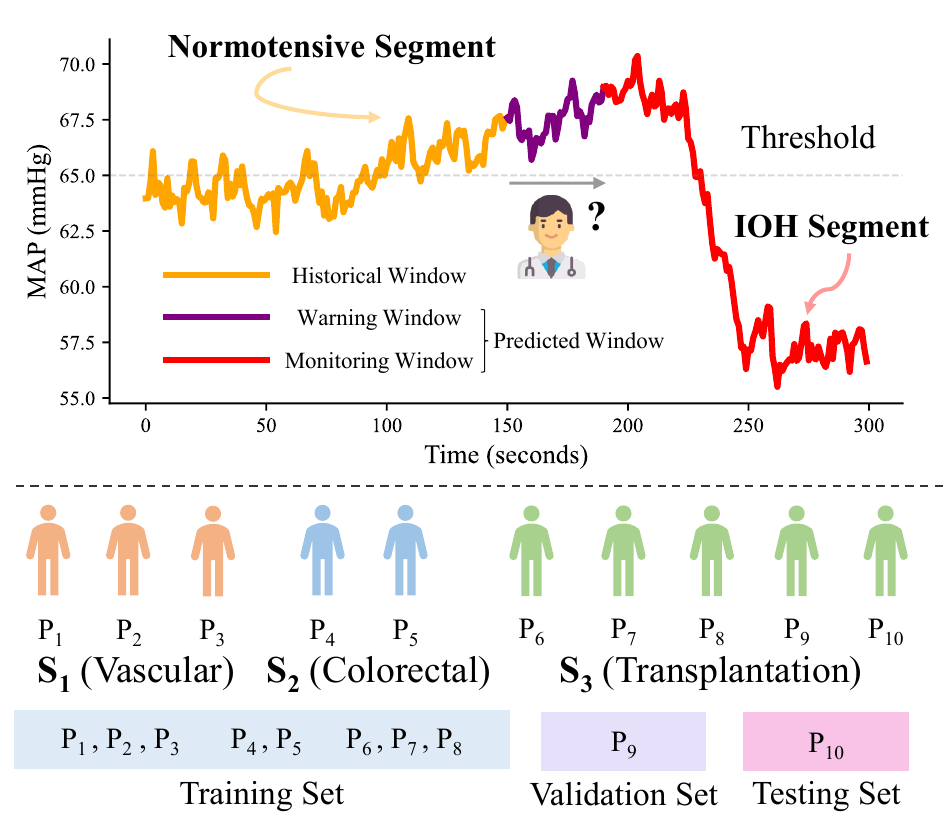}
    \vspace{-0.25in}  
    \caption{Top: Temporal segmentation for IOH prediction. The MAP curve is divided into historical window (orange), warning window (purple), and monitoring window (red). Bottom: Patients are split by procedure into training, validation, and test sets to ensure subject independence and prevent data leakage.}
    \label{premilaries}  
    \vspace{0.1in}
\end{wrapfigure}

\paragraph{Definition of Intraoperative Hypotension.}

Intraoperative hypotension (IOH) is defined according to clinically established thresholds. An IOH event is identified when the mean arterial pressure (MAP) remains below 65 mmHg for at least one continuous minute~\cite{p1, p2}. Systolic blood pressure (SBP) and diastolic blood pressure (DBP) denote the maximum and minimum pressures within a cardiac cycle, respectively. The MAP~\cite{p3}, a critical indicator of cardiac output and systemic vascular resistance~\cite{R10}, is calculated as:

\vspace{-0.1in}
\begin{equation}
\text{MAP} = \frac{\text{SBP} + 2 \times \text{DBP}}{3},
\end{equation}
\vspace{-0.1in}

\paragraph{Series Instance Construction.}
Given a historical window of length \( L \), the model predicts a future MAP series of length \( T \), as illustrated in Fig.~\ref{premilaries}. To prevent label leakage and enable realistic forecasting, instances with historical windows overlapping IOH episodes are excluded. To mitigate class imbalance and capture temporal dynamics, we adopt an adaptive slicing strategy: negative instances are sampled at regular intervals $\Delta_{\text{Normal}}$ to reduce redundancy, while positive instances linked to IOH are sampled more frequently at intervals $\Delta_{\text{IOH}}$ to ensure adequate coverage of critical transitions. 

\vspace{-0.05in}

\paragraph{Surgery Aware Subject Splitting.}
To ensure realistic generalization, we employ a subject independent split by assigning each patient exclusively to the training, validation, or test set, as shown in Fig.~\ref{premilaries}. Patients are grouped according to their surgery type, and each group is assigned to only one data partition. This stratification helps maintain a balanced distribution of surgery types across splits, thus mitigating any distributional shifts caused by surgery specific hypotension risks. This strategy also prevents the memorization of subject-specific patterns and reflects real world deployment scenarios. Moreover, it enables a clear separation of each patient's static attributes and temporal waveform data across splits, which facilitates model generalization to unseen individuals.

\paragraph{IOH Event Evaluation.} 
The ground-truth label for each timestamp is assigned based on whether the subsequent one-minute MAP series remains continuously below 65 mmHg. A predicted IOH event is assigned if more than 60\% of the forecasted MAP values within the same one-minute window fall below this threshold. Model performance is evaluated using pointwise metrics to assess the accuracy of MAP forecasting at each timestep, and instance-level metrics to capture the model’s effectiveness in detecting IOH events.

\paragraph{Problem Definition.}

We define the multimodal dataset as \( \mathcal{X} = \{(a_i, g_i, s_i, \boldsymbol{x}_i) \mid i \in [N]\} \), where each record corresponds to a perioperative patient \( p_i \in \mathcal{P} = \{ p_1, p_2, \ldots, p_N \} \). Here, \( a_i \), \( g_i \), and \( s_i \) denote the patient's age, gender, and type of surgery, respectively, and \( \boldsymbol{x}_i \) represents the associated MAP time series. To support early detection and timely intervention, we formulate IOH prediction as a future MAP forecasting task. For each patient $p_i$, given a MAP historical series \( \boldsymbol{x}_{i,1:l} \) and static attributes \( (a_i, g_i, s_i) \), the objective is to forecast the predicted series \( \boldsymbol{x}_{i,l+1:l+t} \). The predicted series is partitioned into two segments: a two-minute warning window that captures rapid hemodynamic fluctuations for early identification of instability~\cite{p4}, followed by a monitoring window for confirming the occurrence of IOH events.

\vspace{-0.15in}

\section{Methodology}

\vspace{-0.05in}

\begin{figure}
    \centering
    \includegraphics[width=1\linewidth]{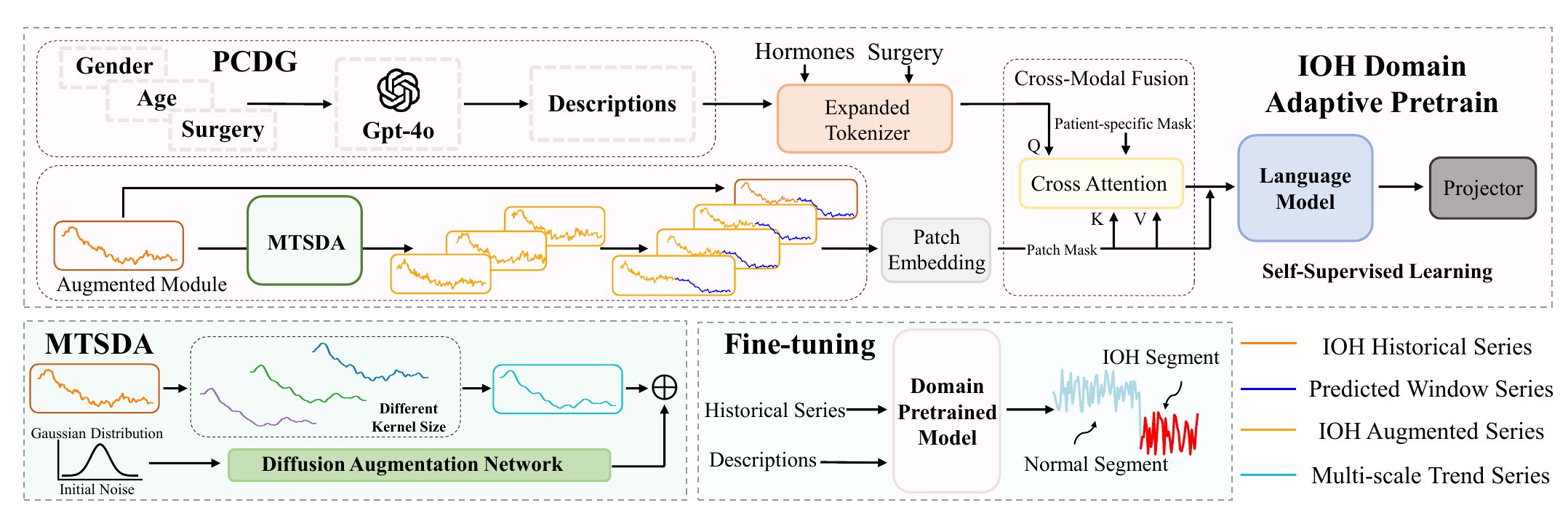}
    \caption{Illustration of our framework. MTRDA decomposes MAP series into multi-scale, and enhances IOH historical series via diffusion-based augmentation. IOH domain adaptive pretraining aligns the augmented IOH series and clinical descriptions through dual-masked cross-attention under a self-supervised objective. Task fine-tuning incorporates labeled normotensive and IOH series with an IOH-specific MSE loss to refine event detection.}
    \label{fig:network}
    \vspace{-0.2in}
\end{figure}

In this section, we describe the proposed multimodal framework IOHFuseLM for intraoperative hypotension (IOH) prediction. As shown in Fig.~\ref{fig:network}, the framework consists of four components: personalized clinical description generation, multi-scale trend-residual diffusion augmentation, domain adaptive pretrain, and task fine-tuning. The model is built on the GPT-2 architecture~\cite{gpt2}.

\vspace{-0.1in}

\subsection{Personalized Clinical Description Generation}

To incorporate static features effectively, we propose a template-guided personalized clinical description generation (\textbf{PCDG}) framework that integrates multiple sources of medical knowledge, including physician recommendations, institutional expertise, and relevant literature. By leveraging hospital-specific data and retrospective studies~\cite{M3, M4}, the framework generates individualized narratives enriched with personalized insights and contextualized by static IOH-related features.

For each patient \( p_i \), the personalized clinical description is defined as:
\begin{equation}
    \boldsymbol{d}_i = \phi(a_i, g_i, s_i),
\end{equation}
Here, \( \phi \) represents GPT-4o that generates clinical descriptions \( \boldsymbol{d}_i \) based on the static attributes \( (a_i, g_i, s_i) \), following a predefined medical template. To enhance the clinical relevance of the language model, the tokenizer is extended with hormone and surgery related terms. The resulting dataset can be formally represented as:
\begin{equation}
    \mathcal{X}_1 = \{ (\boldsymbol{d}_i, \boldsymbol{x}_i ) \mid i \in [N] \}.
\end{equation}

\vspace{-0.1in}

\subsection{Multi-Scale Trend-Residual Diffusion Augmentation}

To alleviate the challenge of scarce IOH cases, which hampers accurate modeling of hemodynamic series, we propose the Multi-Scale Trend Residual Diffusion Augmentation (\textbf{MTRDA}) framework. This approach enhances the representation and generation of sparse MAP series, particularly those containing IOH events. MTRDA improves the model’s ability to learn both broad temporal patterns and fine-grained variations within hypotensive intervals.

Adaptive smoothing is initially performed on the MAP series defined over the historical window to extract underlying trends. Specifically, we employ a set of centered sliding average filters with predefined odd-length window kernels \( \mathcal{S}=\{w_1, w_2, \ldots, w_{|\mathcal{S}|}\} \). For each scale   \( w_s\in \mathcal{S} \), the smoothed series \( \boldsymbol{x}_i^{(s)} \) is computed as:
\vspace{-0.03in}
\begin{equation}
    \boldsymbol{x}_{i,t}^{\langle s\rangle} = \frac{1}{w_s} \sum_{\tau = -\lfloor w_s/2 \rfloor}^{\lfloor w_s/2 \rfloor} \boldsymbol{x}_{i.t+\tau},
\end{equation}
where \( \lfloor \cdot \rfloor \) denotes the floor operator. Boundary values are handled via symmetric padding. The final multiscale trend estimate is obtained by averaging across all scales:
\vspace{-0.03in}
\begin{equation}
    \boldsymbol{x}_{i, \text{trend}} = \frac{1}{\left| \mathcal{S} \right|} \sum_{s=1}^{\left|\mathcal{S}\right|} \boldsymbol{x}^{\langle s \rangle}_i,
\end{equation}

\vspace{-0.08in}
\begin{equation}
    \boldsymbol{x}_{i, \text{residual}} = \boldsymbol{x}_{i,1:l} - \boldsymbol{x}_{i, \text{trend}},
\end{equation}

where \( \boldsymbol{x}_{i, \text{trend}} \) and \( \boldsymbol{x}_{i, \text{residual}} \) represent the trend and residual components of the series, respectively. Short windows capture rapid fluctuations indicative of oscillatory patterns, while long windows reveal sustained trends linked to patient status. This multiscale smoothing strategy preserves structural patterns \( X^{\text{trend}}_i \) across temporal levels and facilitates early IOH detection. The residual component \( \boldsymbol{r}_i \) retains detailed variations reflecting subtle physiological dynamics.

To enhance the residual component \( \boldsymbol{x}_{i, \text{residual}} \) by preserving the overall MAP trend while enriching fine-grained fluctuations, MTRDA incorporates a diffusion-based generative mechanism that learns to reconstruct and refine the residual series through iterative denoising.
\begin{equation}
    \boldsymbol{x}_{i, \text{residual}}^{(k)} = \sqrt{\bar{\alpha}_k} \, \boldsymbol{x}_{i, \text{residual}}^{(0)} + \sqrt{1 - \bar{\alpha}_k} \, \boldsymbol{\epsilon},
\end{equation}
where \( k \) denotes the diffusion step, \( \bar{\alpha}_k \) denotes the cumulative product of the noise schedule coefficients, and \( \boldsymbol{\epsilon} \sim \mathcal{N}(0, \boldsymbol{I}) \) is standard Gaussian noise.
\vspace{-0.1in}

\begin{align}
L_{\text{ELBO}} = \mathbb{E}_{\boldsymbol{x}_{i, \text{residual}}, k} \left\| \boldsymbol{x}_{i, \text{residual}}^{(0)} - f_\theta\left( \boldsymbol{x}^{(k)}_{i, \text{residual}}, k \right) \right\|^2.
\label{denoise_loss}
\end{align}
\vspace{-0.1in}

The diffusion augmentation network \( f_\theta \) comprises three modules. During training, the embedding module encodes the residual series into a high-dimensional latent space using multilayer perceptrons and learnable positional encodings, with diffusion step \( k \) integrated via sinusoidal encodings~\cite{M1,M2} and Adaptive Layer Normalization (AdaLN)~\cite{adaln}. The encoded features are passed to a lightweight denoising decoder composed of stacked linear layers and normalization blocks, which iteratively refine the residual series while reducing computational overhead. A projection layer then maps the refined representation back to the residual space and combines it with the trend component \( \boldsymbol{x}_{i,\text{trend}} \) to generate the output and compute the training loss. For each original series, initial noise is sampled from a Gaussian distribution and passed through the trained network \( f_\theta \) together with the trend component \( \boldsymbol{x}_{i,\text{trend}} \), generating \( H \) augmented MAP series. These are denoted as \( \boldsymbol{X}' = \{ \boldsymbol{x}'^{(1)}, \ldots, \boldsymbol{x}'^{(H)} \} \) and preserve both transient anomalies and subtle fluctuations. We then construct the extended dataset as:
\begin{equation}
\mathcal{X}_2 = \mathcal{X}_1 \cup \left\{ \left( \boldsymbol{d}_i, \boldsymbol{x}_{i,1:l}'^{(j)} \oplus \boldsymbol{x}_{i,l+1:l+t} \right) \mid i \in [N], j \in [H] \right\}.
\end{equation}

This reconstruction process captures fine-grained residual patterns within the historical IOH window, thereby enhancing the fidelity and informativeness of sparse IOH series.

\vspace{-0.1in}

\subsection{Domain Adaptive Pretraining}

Pretraining has demonstrated remarkable effectiveness in time series analysis~\cite{M5}. Our objective is to endow a language model with the capability to identify and comprehend temporal IOH dynamics, while enabling effective cross-modal fusion between physiological time series and patient-specific clinical text. To harness this potential in the context of IOH forecasting, we propose a domain adaptive pretraining strategy that aligns personalized clinical context with IOH physiological patterns.

Specifically, each input pair \( (\hat{\boldsymbol{x}}_i, d_i) \) is sampled from \( \mathcal{X}_2 \). To enable modality alignment, the MAP series \( \hat{\boldsymbol{x}}_i \) is segmented into fixed-length patches of size \( p \), which are linearly projected into MAP patch tokens. A random masking ratio \( R \) is applied to the resulting tokens to enhance representation learning. The corresponding clinical description \( d_i \) is tokenized using expanded language model tokenizer, yielding the text token \( T_i \) with a maximum length of \( \eta \).

To enable selective fusion of patient-specific textual and physiological features, we construct a patient-specific attention mask \( \boldsymbol{M}_i \). Specifically, we define two binary vectors: the first is a vector \( \boldsymbol{1}_{\frac{l+t}{p}} \), whose length matches the number of series tokens, with all elements set to active. The second is a binary vector \( \boldsymbol{m}_i \) of length \( \eta \), corresponding to the text token \( T_i \), where active elements represent valid tokens and inactive elements represent padding positions. Additionally, we define an all-active vector \( \boldsymbol{1}_\eta \) of length \( \eta \). These masks are combined using elementwise logical conjunction to form the joint attention mask \( \boldsymbol{M}_i \), defined as:

\vspace{-0.15in}
\begin{equation}
    \boldsymbol{M}_i = \boldsymbol{1}_{\frac{l+t}{p}} \cdot \left( \boldsymbol{1}_{\eta} - \boldsymbol{m}_i \right)^\top
\end{equation}

\vspace{-0.15in}
\begin{equation}
    \text{Attention}(\boldsymbol{Q}, \boldsymbol{K}, \boldsymbol{V}) = \text{softmax} \left( \frac{\boldsymbol{Q}\boldsymbol{K}^\top}{\sqrt{d}} - \lambda \cdot \boldsymbol{M}_i \right) \boldsymbol{V},
\end{equation}
where \( \lambda \) is a large constant to suppress attention to semantically misaligned regions. We adopt the token level alignment mechanism~\cite{M6}, which aligns the text token \( T_i \) with the patch token of series \( \hat{\boldsymbol{x}}_i \). The query matrix \( Q \) is derived from the tokenized and projected text tokens \( T_i \), while the key and value matrices \( K \) and \( V \) are obtained from the corresponding MAP series tokens. The resulting representations are concatenated with the series tokens and processed by a pretrained language model, followed by a projection layer. The model is optimized to minimize the mean squared error (MSE) on the masked positions of the time-series tokens.

This pretraining strategy enables the model to learn semantically meaningful interactions between the IOH related MAP series and the corresponding static patient features. It also facilitates modality alignment between IOH series and patient specific information, thereby providing a stronger language model foundation for subsequent hypotension prediction.

\vspace{-0.1in}

\subsection{Task Fine-tuning}

To adapt the pretrained model to the downstream IOH prediction task, the task fine-tuning stage further refines the representations learned during domain adaptive pretraining, thereby enhancing the model's ability to distinguish IOH physiological patterns from normotensive fluctuations.

Each input pair \( (d_i, \boldsymbol{x}_i) \in \mathcal{X}_1 \) is first processed to derive token level representations that capture instance-specific semantic and physiological characteristics. These representations are integrated with the corresponding series embeddings and then passed to the pretrained model. During task fine-tuning, the series embedding and output projection layers are reinitialized for task adaptation, while all other parameters are initialized from the domain adaptive pretraining stage and jointly optimized to improve temporal sensitivity to IOH dynamics.

To enhance sensitivity to IOH abnormalities, an additional loss term is introduced for the timestamps of IOH series, defined as the MSE on those timestamps and weighted by a hyperparameter \( \rho \). The total IOH loss function is given by:
\begin{equation}
\text{Loss} = \text{MSE}_{\text{normal}} + \rho \cdot \text{MSE}_{\text{IOH}},
\end{equation}
where \( \text{MSE}_{\text{normal}} \) and \( \text{MSE}_{\text{IOH}} \) represent the mean squared errors computed over normotensive and hypotensive series, respectively. This task fine-tuning strategy encourages the model to attend to subtle temporal variations indicative of IOH, thereby enhancing predictive performance and facilitating timely clinical intervention.

\vspace{-0.15in}
\section{Experiments}

\subsection{Experimental Setup}
\paragraph{Datasets.}  
We use two clinical intraoperative hypotension (IOH) datasets collected in real-world surgical settings.
\textbf{Clinical IOH Dataset.} The dataset includes intraoperative records from 6,822 patients, featuring MAP time series resampled at 6 and 10 seconds from arterial blood pressure waveforms, together with patient attributes such as age, gender, and surgery type. A total of 1,452 high-quality recordings were retained after preprocessing.
\textbf{VitalDB Dataset.} This dataset originally consisted of 6,388 ABP recordings. After filtering out low-quality samples, 1,522 recordings were retained for downstream experiments.
Both datasets are split into training, validation, and test sets in a 3:1:1 ratio with temporal consistency. We use a 15-minute historical window and predicted horizons of 5, 10, and 15 minutes, guided by clinical evidence on IOH predictability~\cite{E1}. Detailed preprocessing steps and dataset statistics are provided in Appendix~\ref{Dataset Details}.

\paragraph{Baselines.}
We compare our method against six representative time series forecasting models. These include the MLP-based DLinear~\cite{DLinear}, the Transformer-based PatchTST~\cite{PatchTST}, and the frequency domain enhanced Fredformer~\cite{Fredformer}. We also include HMF~\cite{I23}, a model specifically designed for intraoperative hypotension prediction, along with two language model-based approaches: GPT4TS~\cite{GPT4TS}, built on GPT-2, and TimeLLM~\cite{Timellm}, based on LLaMA2-7B~\cite{llama}.

\paragraph{Implementation.}
To assess IOH prediction performance, we report Mean Squared Error (MSE) and Mean Absolute Error (MAE) on hypotensive timestamps. Discriminative ability is measured by the Area Under the ROC Curve (AUC), and Recall reflects early warning effectiveness. All models are trained with the Adam optimizer~\cite{E2}, using a batch size of 8 and an initial learning rate of 0.0001 with a decay factor of 0.75. Experiments are conducted on a NVIDIA RTX 4090 GPU and a server equipped with eight Tesla A100 GPUs. To ensure result reliability, we report averages over three independent runs. Full training configurations are provided in Appendix~\ref{Experiment Details}.

\subsection{Results and Discussion} 

\paragraph{Main Results.} 
We conduct comprehensive experiments on the Clinical IOH dataset and VitalDB dataset. Results are summarized in Table~\ref{tab:main_results}. Experimental results highlight key differences among baseline models. DLinear’s moderate performance reflects limitations in capturing complex temporal dynamics using simple linear decomposition. PatchTST excels in recall and AUC by segmenting series into semantically meaningful patches. Fredformer improves performance by reducing frequency bias but struggles with the high variability of IOH events. HMF extracts temporal features using sliding windows but lacks semantic modeling of baseline blood pressure associated with surgery type, leading to poor generalization across procedures. GPT4TS performs well on high-frequency data, capturing short physiological trends effectively. TimeLLM’s fixed parameters limit its adaptability to distribution shifts in MAP series. IOHFuseLM outperforms others in sparse and high-variability settings by aligning static patient attributes with MAP series and augmenting sparse data with realistic signals, effectively addressing IOH sparsity and variability.

Performance varies distinctly across datasets and sampling rates. VitalDB, with higher IOH event density, generally yields better metrics, particularly for IOHFuseLM, which excels at fine granularity, demonstrating strong temporal pattern extraction capabilities. Clinical IOH, characterized by sparser events, presents greater modeling challenges, yet IOHFuseLM consistently maintains strong performance across coarser sampling intervals. These results emphasize adaptability and the effectiveness of its multimodal context integration and data augmentation strategies.

\begin{table}[htbp]
  \centering
  \small
  \caption{Performance comparison of different models on the Clinical IOH and VitalDB datasets under varying sampling rates. The best result for each metric is indicated in bold.}
  \setlength{\tabcolsep}{5pt}
  \renewcommand\arraystretch{1.1}
  \resizebox{0.95\textwidth}{!}{
  \begin{tabular}{lcclcccc}
    \toprule
    Dataset & Historical Window & Sampling (s) & Model & MSE$_\mathrm{IOH}$ & MAE$_\mathrm{IOH}$ & Recall (\%) & AUC \\
    \midrule
    \multirow{14}{*}{Clinical IOH} & \multirow{7}{*}{150} & \multirow{7}{*}{6}
      & DLinear       & 178.6592 & 10.9190 & 36.61\% & 0.6406 \\
      & &               & PatchTST      & 122.2292 & 8.5687 & 63.09\% & 0.6948 \\
      & &               & Fredformer    & 99.0389  & 7.7170 & 59.98\% & 0.6985 \\
      & &                & HMF           & 114.6592 & 8.3079 & 50.76\% & 0.6737 \\
      & &                & GPT4TS        & 119.0686 & 8.4467 & 59.37\% & 0.6991 \\
      & &                & TimeLLM       & 133.2503 & 9.2422 & 46.58\% & 0.6687 \\
      & &                & \textbf{IOHFuseLM} & \textbf{88.9192} & \textbf{7.4921} & \textbf{74.00\%} & \textbf{0.7130} \\
    \cmidrule{2-8}
     & \multirow{7}{*}{90} & \multirow{7}{*}{10}
      & DLinear       & 127.0857 & 8.6489 & 51.49\% & 0.6933 \\
      & &                & PatchTST      & 125.5609 & 8.8750 & 56.84\% & 0.7044 \\
      & &                & Fredformer    & 103.5407 & 8.0945 & 53.17\% & 0.6935 \\
      & &                & HMF           & 121.1721 & 8.7806 & 51.40\% & 0.6853 \\
      & &                & GPT4TS        & 91.4255  & 7.4369 & 62.68\% & 0.7309 \\
      & &                & TimeLLM       & 118.6838 & 8.7864 & 50.45\% & 0.6913 \\
      & &                & \textbf{IOHFuseLM} & \textbf{87.6147} & \textbf{7.3933} & \textbf{74.46\%} & \textbf{0.7425} \\
    \midrule
    \multirow{7}{*}{VitalDB} & \multirow{7}{*}{3}
     & \multirow{7}{*}{300} & DLinear       & 92.2800  & 7.3100 & 33.48\% & 0.6300 \\
      & &                & PatchTST      & 99.3965  & 7.6942 & 52.62\% & 0.6443 \\
      & &                & Fredformer    & 69.5776  & 6.0244 & 49.94\% & 0.6640 \\
      & &                & HMF           & 76.5757  & 6.5151 & 49.73\% & 0.6501 \\
      & &                & GPT4TS        & 61.7742  & 5.5824 & 57.39\% & 0.6885 \\
      & &                & TimeLLM       & 82.3817  & 7.0517 & 30.36\% & 0.6068 \\
      & &                & \textbf{IOHFuseLM} & \textbf{58.3511} & \textbf{5.1251} & \textbf{70.10\%} & \textbf{0.7086} \\
    \bottomrule
  \end{tabular}
  }
  \label{tab:main_results}
  \vspace{-0.1in}
\end{table}

\paragraph{Ablation Results.} 
The following ablated variants are evaluated: 

\begin{itemize}[leftmargin=*, itemsep=2pt]
    \item \textbf{IOHFuseLM\textsuperscript{1}}: Excludes the clinical description \( d_i \), only modeling the MAP time series.
    \item \textbf{IOHFuseLM\textsuperscript{2}}: Utilizes the original GPT-2 tokenizer without vocabulary expansion.
    \item \textbf{IOHFuseLM\textsuperscript{3}}: Conducts domaim adaptive pretraining exclusively on the original dataset \( \mathcal{X}_1 \), omitting any diffusion-based augmentation.
    \item \textbf{IOHFuseLM\textsuperscript{4}}: Removes the domaim adaptive pretraining stage and directly applies task fine-tuning on the downstream IOH prediction task.
\end{itemize}

We conduct a detailed ablation study on the Clinical IOH dataset sampled at 10-second intervals, using a historical window of 15 minutes and predicted horizons of 5, 10, and 15 minutes. Table~\ref{tab:ablation} summarizes the impact of removing each key component from IOHFuseLM. The results confirm that every component is essential for addressing the challenges discussed above. Removing static attributes degrades performance by eliminating personalized priors that help identify MAP trends and variability, reducing the ability to detect abnormal patterns and generalize across populations and surgery types. Excluding the expanded tokenizer weakens the model’s ability to associate clinical terminology with physiological patterns, diminishing cross-modal representation learning. Using only the original dataset \( \mathcal{X}_1 \) for domain adaptation pretraining, instead of the MTRDA dataset, reduces sensitivity to rare and short-duration IOH episodes, demonstrating the benefit of synthetic variability in mitigating data sparsity. Finally, omitting the domain adaptation pretraining stage leads to consistent performance degradation across all metrics, confirming that prior exposure to IOH patterns enhances generalization under limited supervision.

\begin{table}[htbp]
  \centering
  \vspace{-0.1in}
  \caption{Ablation study of model components on the Clinical IOH dataset.}
  \small
  \setlength{\tabcolsep}{8pt}
  \renewcommand{\arraystretch}{1.2}
  \resizebox{0.95\textwidth}{!}{
  \begin{tabular*}{\linewidth}{@{\extracolsep{\fill}}llcccc}
    \toprule
    Dataset & Model Variant & MSE$_\mathrm{IOH}$ & MAE$_\mathrm{IOH}$ & Recall & AUC \\
    \midrule
    \multirow{5}{*}{Clinical IOH} & IOHFuseLM                   & \textbf{87.6147} & \textbf{7.3933} & \textbf{74.46\%} & \textbf{0.7425} \\
                                  & IOHFuseLM\textsuperscript{1} & 87.6824          & 7.4604          & 68.62\%          & 0.7215 \\
                                  & IOHFuseLM\textsuperscript{2} & 95.4979          & 7.8967          & 69.42\%          & 0.7287 \\
                                  & IOHFuseLM\textsuperscript{3} & 107.7359         & 8.3523          & 67.78\%          & 0.7213 \\
                                  & IOHFuseLM\textsuperscript{4} & 98.1118          & 7.9061          & 67.85\%          & 0.7192 \\
    \bottomrule
  \end{tabular*}}
  \label{tab:ablation}
  \vspace{-0.1in}
\end{table}

\paragraph{Transfer Results.} 

To evaluate the model’s generalization ability, we conducted transfer learning experiments on a newly curated cohort of patients with surgical durations between 600 and 1000 seconds, sampled every 6 seconds. The historical and predicted windows were set to 3 and 5 minutes, respectively. The model was pretrained on the Clinical IOH dataset and tested under the settings, focusing on adult patients aged 18–65 years. As shown in Table~\ref{tab:transfer_experiment_results}, transfer learning substantially improved performance across all metrics, particularly in recall and AUC. These gains indicate enhanced sensitivity to IOH events and better discrimination under demographic and procedural variability. The results highlight the effectiveness of domaim adaptive pretraining and personalized context integration in enabling effective generalization across diverse clinical scenarios.

\begin{table}[htbp]
  \centering
  \vspace{-0.1in}
  \caption{Performance comparison with and without transfer learning on the Clinical IOH dataset.}
  \vspace{0.5em}
  \small
  \renewcommand\arraystretch{1.1}
  \setlength{\tabcolsep}{4pt}
  \resizebox{\linewidth}{!}{
    \begin{tabular}{lccccccc}
      \toprule
      Transfer Setting & Historical Window & Predicted Window & MSE$_\mathrm{IOH}$ & MAE$_\mathrm{IOH}$ & Recall & AUC \\
      \midrule
      Without Transfer Learning & 30 & 50 & 216.4261 & 14.2554 & 0.00\%  & 0.5000 \\
      Transfer Learning         & 30  &   50   & 97.9322  & 9.0675  & 17.65\% & 0.5805 \\
      \bottomrule
    \end{tabular}
  }
  \label{tab:transfer_experiment_results}
  \vspace{-0.2in}
\end{table}

\begin{figure}[h]
    \centering
    \vspace{-0.1in}
    \includegraphics[width=\linewidth]{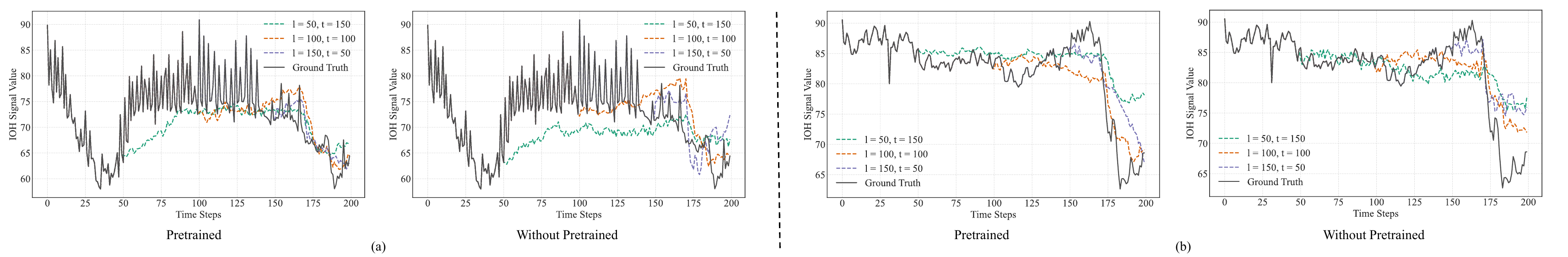}
    \caption{Qualitative comparison of models with and without domaim adaptive pretraining under two representative IOH patterns: (a) gradually declining MAP; (b) rapidly declining MAP.}
    \vspace{-0.2in}
    \label{fig:Visual_pretrain}
    \vspace{-0.05in}
\end{figure}

\paragraph{Visual Evidence for the Impact of Domaim Adaptive Pretraining.} 
To qualitatively evaluate the effect of domain adaptive pretraining, we visualize MAP forecasts under two representative IOH patterns: one with a gradual decline and the other with a rapid decline. As shown in Figure~\ref{fig:Visual_pretrain}, we compare models trained with and without pretraining using the same total series length \( l+t \), where the \( l \) are set to 50 ,100 and 150, respectively. In both patterns, the models with pretraining produce predictions that more closely follow the ground truth, especially in their ability to capture downward trends in blood pressure. This advantage is particularly clear in the rapid-decline scenario, where models without pretraining tend to respond more slowly and deviate further from actual MAP values. When the historical length \( l \) is sufficiently long, for example 150, the pretrained model produces stable and accurate forecasts, benefiting from richer temporal context and more reliable trend estimation. This suggests that the proposed pretraining method enhances the model’s sensitivity to temporal changes and improves its ability to recognize IOH-specific patterns.

\begin{figure}[h]
    \centering
    \vspace{-0.1in}
    \includegraphics[width=\linewidth]{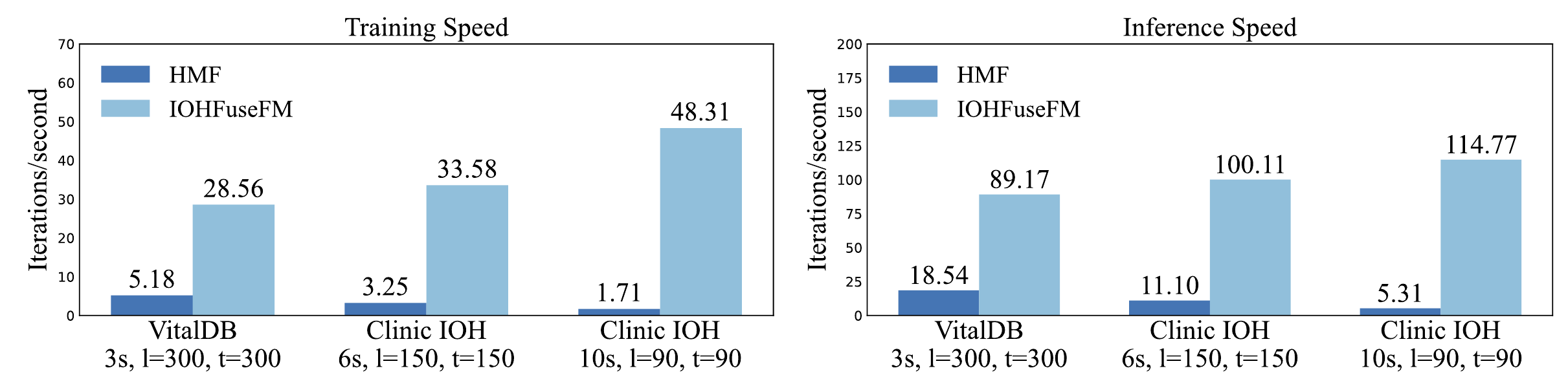}
    \caption{
        Comparison of training and inference speed between IOHFuseLM and HMF.
    }
    \label{fig:Speed_Com}
    \vspace{-0.15in}
\end{figure}

\paragraph{Deployability Evaluation.}
\label{Deployability Evaluation}
To evaluate the practical deployability of our framework, we compare its training and inference efficiency with HMF~\cite{I23}, a baseline for IOH prediction. As shown in Figure~\ref{fig:Speed_Com}, IOHFuseLM consistently achieves higher runtime efficiency across configurations from the VitalDB and Clinical IOH datasets. The improvements in both training and inference highlight the model’s computational advantage and its suitability for deployment in clinical environments where timely response is critical.

\section{Conclusion}
In this work, we introduced IOHFuseLM, a multimodal framework for sparse intraoperative hypotension (IOH) prediction that integrated static patient attributes with dynamic physiological time series data. Evaluations on two real world intraoperative datasets showed that IOHFuseLM consistently outperformed competitive baselines, with particularly strong performance under coarse sampling and sparse event conditions. The proposed approach achieved higher recall and area under the curve (AUC) by effectively capturing patient-specific variability and extracting features with rich temporal and semantic content. Compared to previous methods that relied solely on physiological series or used simple feature concatenation, IOHFuseLM demonstrated better flexibility and generalization through data augmentation during pretraining and structured integration of patient attributes. These results suggest that the model can support personalized and timely clinical monitoring, which may help facilitate early intervention, improve hemodynamic stability during surgery, and reduce the risk of postoperative complications.

\clearpage

\bibliographystyle{unsrt}
\bibliography{neurips_2025}

\medskip

{
\small



\clearpage

\appendix

\section{Dataset Details}
\label{Dataset Details}

\begin{table}[htbp]
  \centering
  \scriptsize
  \caption{Statistics of the VitalDB and Clinical IOH datasets under varying sampling settings.}
  \label{tab:dataset_stats}
  \renewcommand{\arraystretch}{1.1}
  \setlength{\tabcolsep}{3pt}
  \resizebox{0.85\linewidth}{!}{
  \begin{tabular}{lccccccccc}
    \toprule
    Dataset & Sampling(s) & Historical Window & Predicted Window & Train & Val & Test & Surgeries & N & IOH$_\text{train}$ \\
    \midrule
    \multirow{6}{*}{Clinic-IOH} 
        & \multirow{3}{*}{6} & \multirow{3}{*}{150} & 50  & 2749  & 852   & 884 & \multirow{6}{*}{28}  & \multirow{6}{*}{1452} & \multirow{6}{*}{135} \\
    & & & 100 & 3348  & 937   & 1138 & & & \\
    & & & 150 & 4507  & 1578  & 1804 & & & \\
    \cmidrule{2-7}
    & \multirow{3}{*}{10} & \multirow{3}{*}{90} & 30  & 10765 & 3236 & 3481 & & & \\
    & & & 60  & 12334 & 3752 & 4227 & & & \\
    & & & 90  & 13320 & 4166 & 4609 & & & \\
    \midrule
    \multirow{3}{*}{VitalDB}
        & \multirow{3}{*}{3} & \multirow{3}{*}{300} & 100 & 18172 & 2308 & 2299 & \multirow{3}{*}{12} & \multirow{3}{*}{1522} & \multirow{3}{*}{2026} \\
    & & & 200 & 27031 & 3635 & 3356 & & & \\
    & & & 300 & 38479 & 5201 & 4647 & & & \\
    \bottomrule
  \end{tabular}
  }
\end{table}

Table~\ref{tab:dataset_stats} summarizes the key statistics of the two intraoperative hypotension (IOH) datasets used in this study: Clinical IOH and VitalDB. For each dataset, we list the sampling frequency, historical historical length \( l \), predicted horizon \( t \), and the number of training, validation, and test samples. The table also includes the number of unique patients (\( N \)) and surgeries, along with the number of IOH events identified in the training set (IOH\(_\text{train}\)) according to our clinical threshold definition.
\textbf{Clinical IOH Dataset.} This dataset consists of intraoperative data collected from 6,822 patients undergoing anesthesia. It contains high-resolution arterial blood pressure (ABP) waveforms sampled at 100\,Hz and structured patient information including age, gender, and surgery type. To accommodate different temporal resolutions, ABP waveforms were processed into MAP series and resampled at 6,s and 10,s intervals. Segments shorter than 1,000 seconds were discarded, yielding 1,452 valid recordings for analysis. Data acquisition was approved by the institutional ethics committee.
\textbf{VitalDB Dataset.} Derived from the public VitalDB repository, this dataset initially included 6,388 intraoperative recordings with ABP values sampled every 3 seconds. We excluded recordings with more than 20\% missing data in the observation window. After filtering, 1,522 high-quality samples remained for downstream tasks.
\textbf{Data Splitting and Forecasting Settings.} Both datasets are split into training, validation, and test subsets using a 3:1:1 ratio, preserving temporal consistency without shuffling. Each model ingests a fixed 15-minute historical MAP window and predicts MAP trajectories over future horizons of 5, 10, or 15 minutes. These predicted lengths are chosen based on prior clinical research demonstrating their practical relevance for IOH risk forecasting~\cite{E1}.
The Clinical IOH dataset was de-identified under the HIPAA Safe Harbor method by removing all 18 identifiers. The VitalDB dataset is publicly available under the CC0 1.0 license, allowing unrestricted use for research.

\begin{table}[htbp]
  \centering
  \caption{Hyperparameter configurations across different datasets and settings.}
  \label{tab:hyperparam_settings}
  \renewcommand{\arraystretch}{1.1}
  \setlength{\tabcolsep}{3pt}
  \scriptsize
  \resizebox{0.95\linewidth}{!}{
  \begin{tabular}{lcccccccc}
    \toprule
    Dataset & Sampling(s) & Predicted Window & Pretrain LR & Finetune LR & H & E & \(\Delta_{\text{Normal}}\) & \(\Delta_{\text{IOH}}\) \\
    \midrule
    \multirow{3}{*}{CH-OBPB} 
        & \multirow{3}{*}{6} & 50  & \(10^{-4}\) & \(10^{-5}\) & 4 & 5 & \multirow{3}{*}{150} & \multirow{3}{*}{2} \\
        & & 100 & \(10^{-5}\) & \(5 * 10^{-5}\) & 4 & 5 &                      &                     \\
        & & 150 & \(10^{-4}\) & \(10^{-4}\) & 3 & 2 &                      &                     \\
    \midrule
    \multirow{3}{*}{CH-OBPB} 
        & \multirow{3}{*}{10} & 30  & \(10^{-4}\) & \(10^{-4}\) & 5 & 2 & \multirow{3}{*}{20} & \multirow{3}{*}{1} \\
        & & 60  & \(10^{-5}\) & \(5 * 10^{-5}\) & 4 & 4 &                     &                    \\
        & & 90  & \(5 * 10^{-5}\) & \(10^{-4}\) & 1 & 5 &                     &                    \\
    \midrule
    \multirow{3}{*}{VitalDB} 
        & \multirow{3}{*}{3} & 100 & \(3 * 10^{-5}\) & \(10^{-4}\) & 3 & 2 & \multirow{3}{*}{150} & \multirow{3}{*}{10} \\
        & & 200 & \(5 * 10^{-5}\) & \(10^{-4}\) & 4 & 2 &                     &                     \\
        & & 300 & \(10^{-5}\) & \(10^{-4}\) & 5 & 2 &                     &                     \\
    \bottomrule
  \end{tabular}
  }
\end{table}

\section{Experiment Details}
\label{Experiment Details}

Table~\ref{tab:hyperparam_settings} summarizes the hyperparameter configurations used across different datasets and experimental settings. Specifically, it includes the predicted window length \( t \), learning rates for domain adaptive pretraining and task fine-tuning, the number of augmented series \( H \), the GPT Layers \( E \), and the sampled intervals \( \Delta_{\text{Normal}} \) and \( \Delta_{\text{IOH}} \). The batch size is fixed at 4 for pretraining and 8 for fine-tuning. The pretraining masking ratio \( R \) is set to 0.2, and the hyperparameter \( \rho \) of IOH loss is set to 10. The diffusion process utilizes $K = 50$ steps with a cosine variance schedule~\cite{Timegrad} from $\beta_1 = 10^{-4}$ to $\beta_K = 0.5$. Most baseline models, including our proposed method, are evaluated on a single NVIDIA RTX 4090 GPU to ensure a fair comparison under practical deployment settings. However, due to the substantial memory requirements of TimeLLM, which is based on the LLaMA 7B language model, both training and inference for this model are conducted on a server equipped with NVIDIA A100 GPUs. We acknowledge the discrepancy in hardware and note that TimeLLM cannot be executed on the RTX 4090 due to out-of-memory limitations, making the A100 the minimal viable hardware configuration for its evaluation. Additionally, the generation of clinical descriptions in the PCDG module is performed using GPT-4o.

\section{Visualization}
\label{Visualization}

\subsection{Model Prediction Visualization}

\begin{figure}[h]
    \centering
    \includegraphics[width=\linewidth]{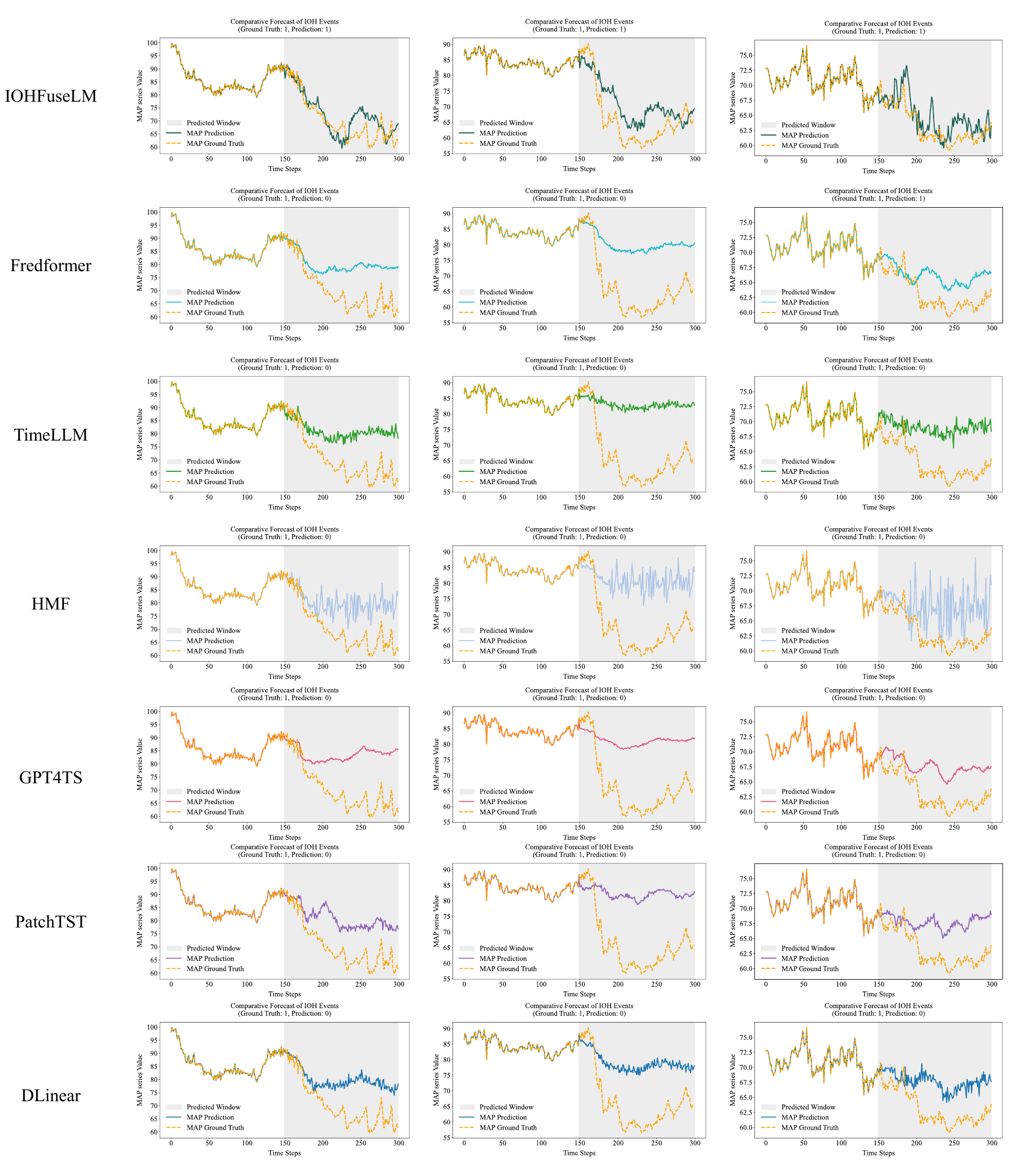}
    \caption{Visual comparison of MAP prediction results across different models.}
    \label{fig:Visual}
\end{figure}

Figure~\ref{fig:Visual} presents a visual comparison of seven models under the 6-second sampling granularity, with a historical window length \( l = 150 \) and a predicted horizon \( t = 150 \). Each row corresponds to one model and each column represents a distinct IOH case. It can be observed that our proposed model consistently identifies hypotensive risks across all three representative IOH events, demonstrating both precise forecasting accuracy and effective event discrimination. In contrast, among the baseline models, only Fredformer correctly identifies the third IOH event, while the others fail to capture the hypotensive onset in this scenario.

\subsection{Augmentation Visualization}

Figure~\ref{fig:sampling_comparison} presents representative examples of MAP time series augmented by the proposed MTRDA framework under two different sampling frequencies. The augmented series preserve the trends extracted through multiscale smoothing and simultaneously introduce fine-grained variations that enrich the temporal structure of the original series. In particular, the augmented outputs retain the essential characteristics of hypotensive episodes while reducing noise, reflecting the ability of MTRDA to reconstruct physiologically meaningful patterns through trend-residual decomposition and diffusion-based enhancement. These results confirm the effectiveness of MTRDA in improving the representation quality of sparse IOH series under varying temporal resolutions.

\begin{figure}[h]
  \centering
  \begin{subfigure}[b]{0.32\textwidth}
    \includegraphics[width=\linewidth]{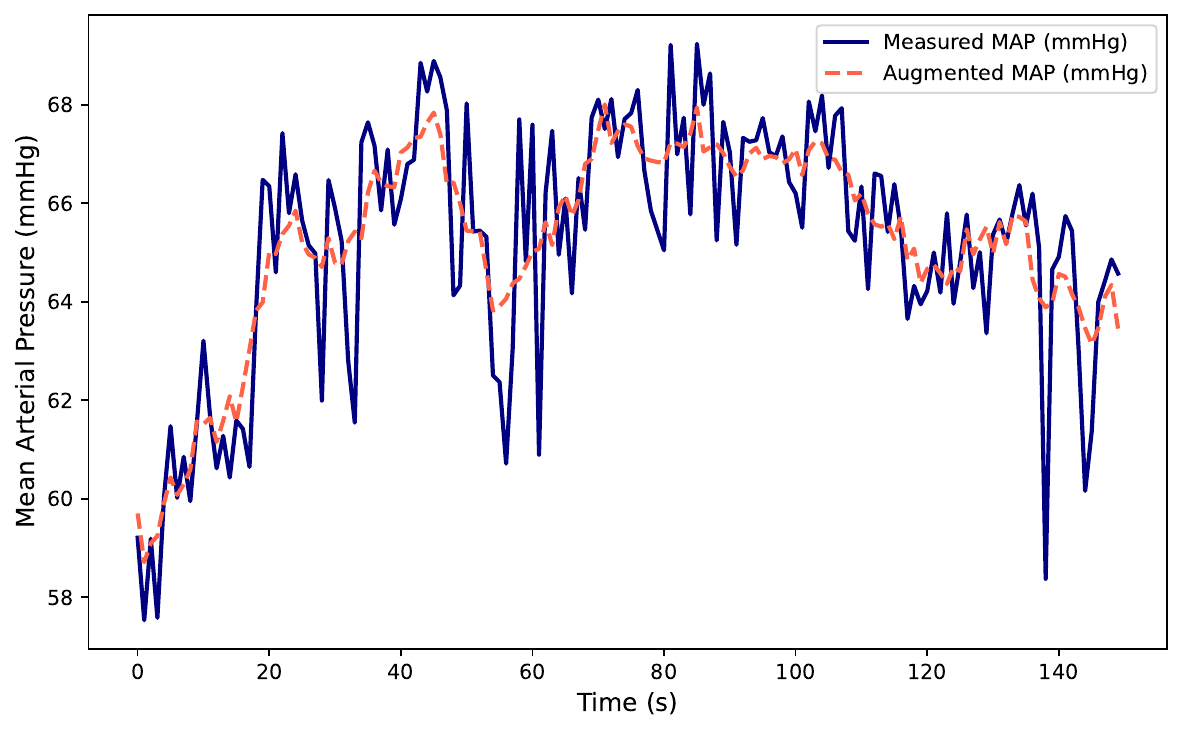}
    \caption{Sample 1 (6s)}
  \end{subfigure}
  \hfill
  \begin{subfigure}[b]{0.32\textwidth}
    \includegraphics[width=\linewidth]{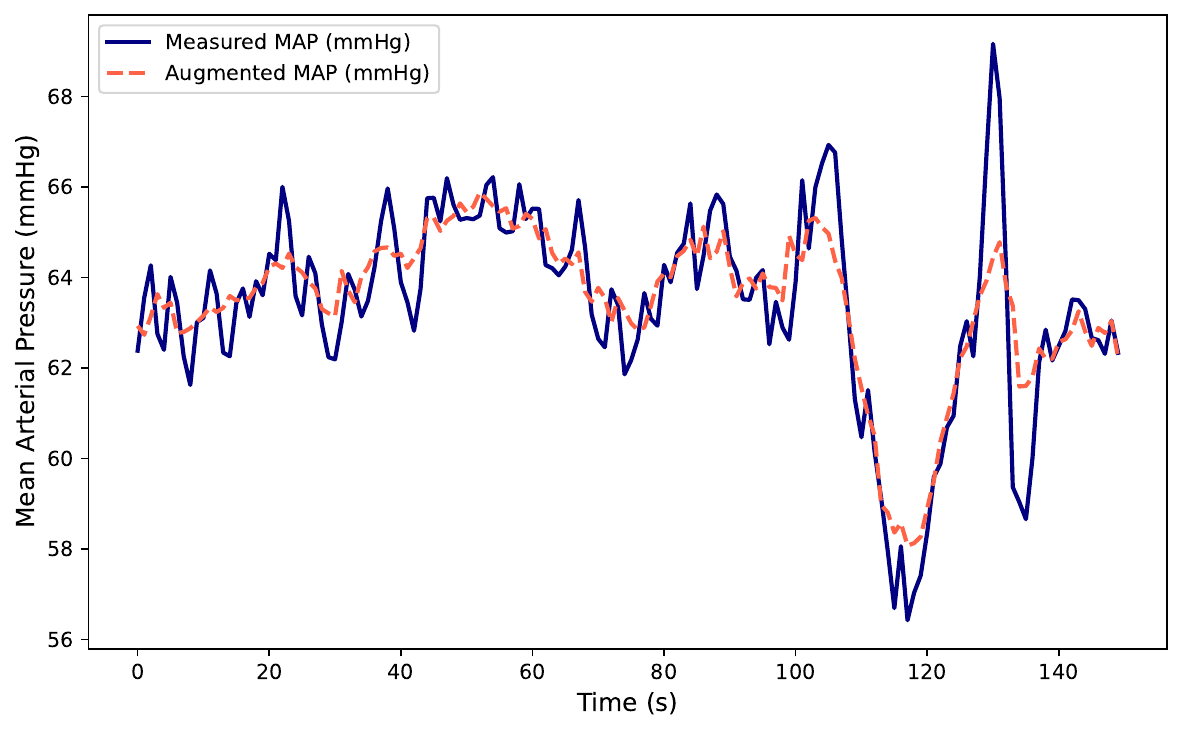}
    \caption{Sample 2 (6s)}
  \end{subfigure}
  \hfill
  \begin{subfigure}[b]{0.32\textwidth}
    \includegraphics[width=\linewidth]{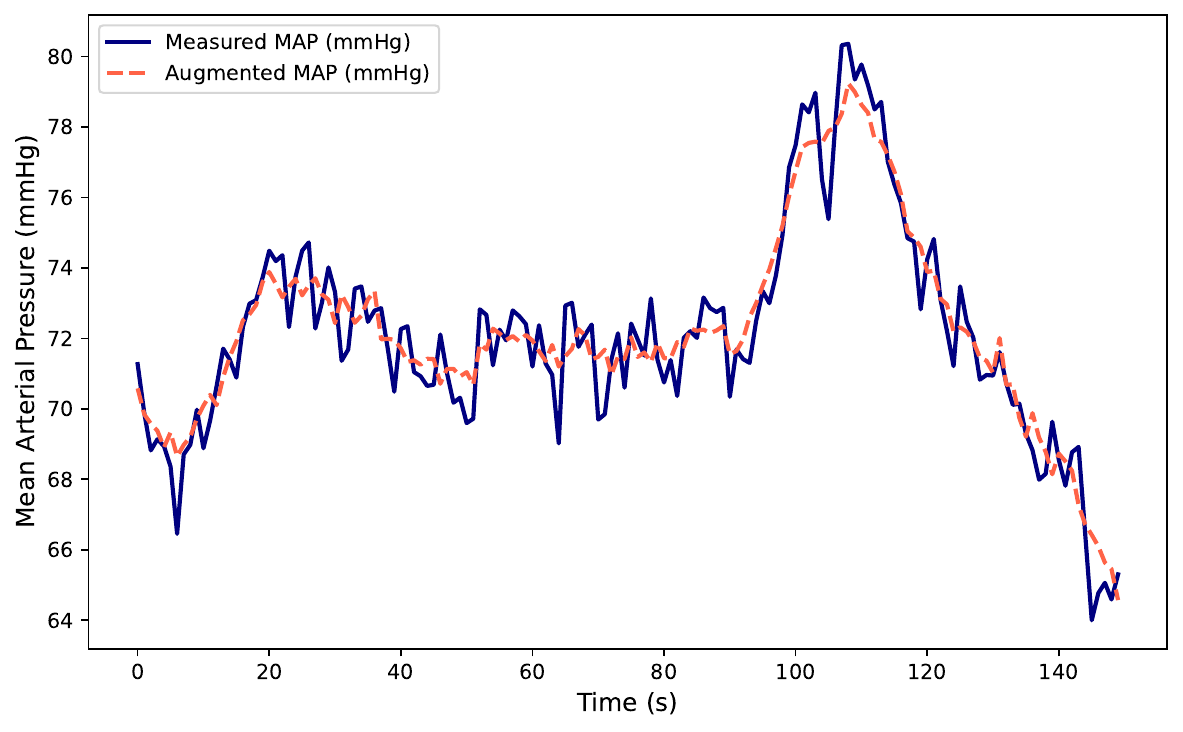}
    \caption{Sample 3 (6s)}
  \end{subfigure}

  \vspace{0.8em}  

  \begin{subfigure}[b]{0.32\textwidth}
    \includegraphics[width=\linewidth]{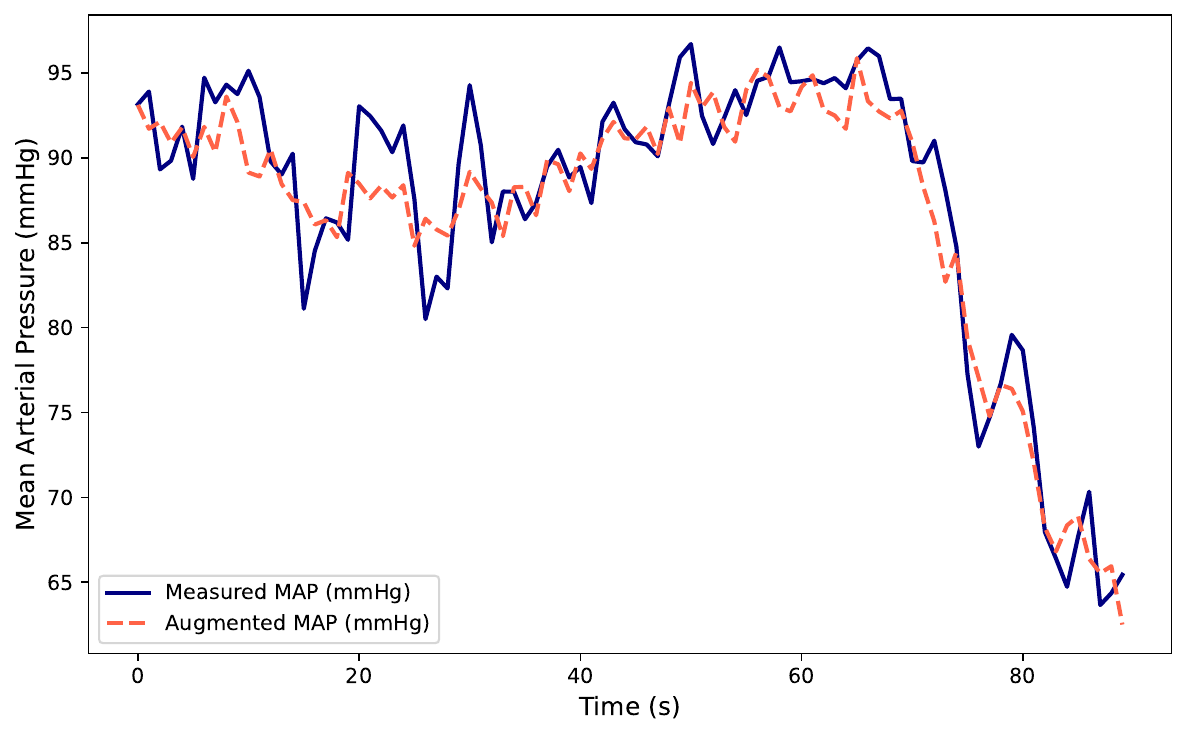}
    \caption{Sample 1 (10s)}
  \end{subfigure}
  \hfill
  \begin{subfigure}[b]{0.32\textwidth}
    \includegraphics[width=\linewidth]{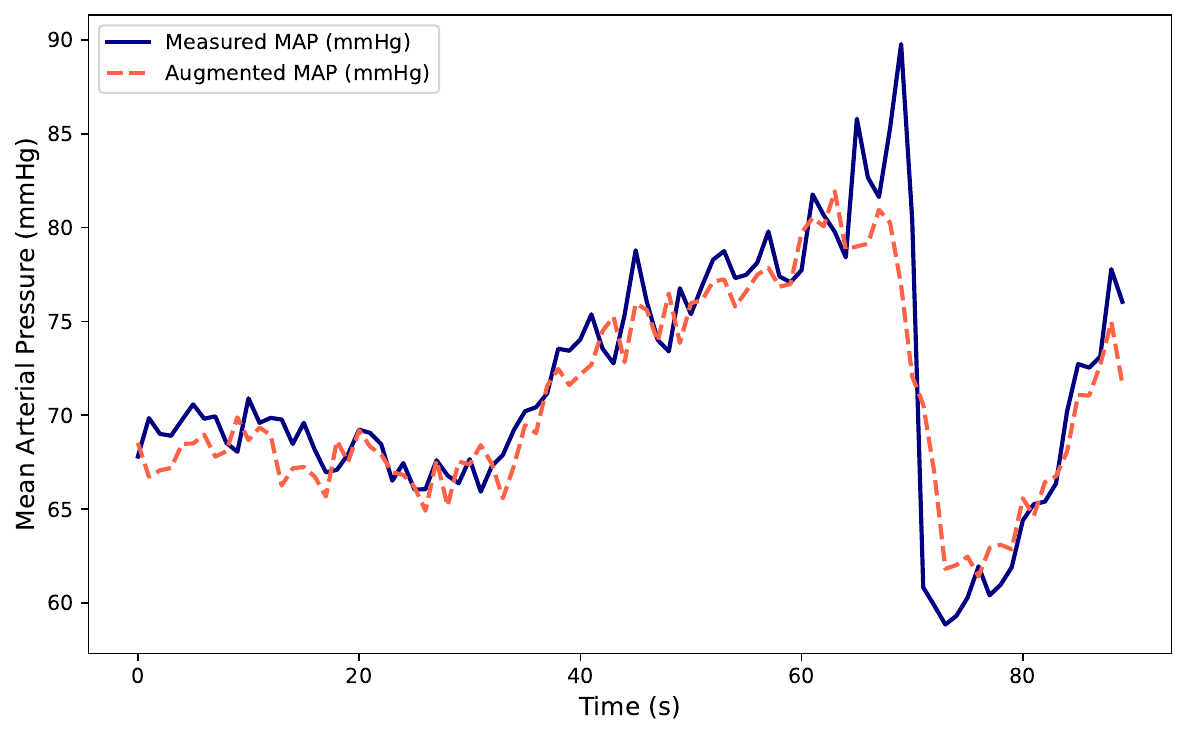}
    \caption{Sample 2 (10s)}
  \end{subfigure}
  \hfill
  \begin{subfigure}[b]{0.32\textwidth}
    \includegraphics[width=\linewidth]{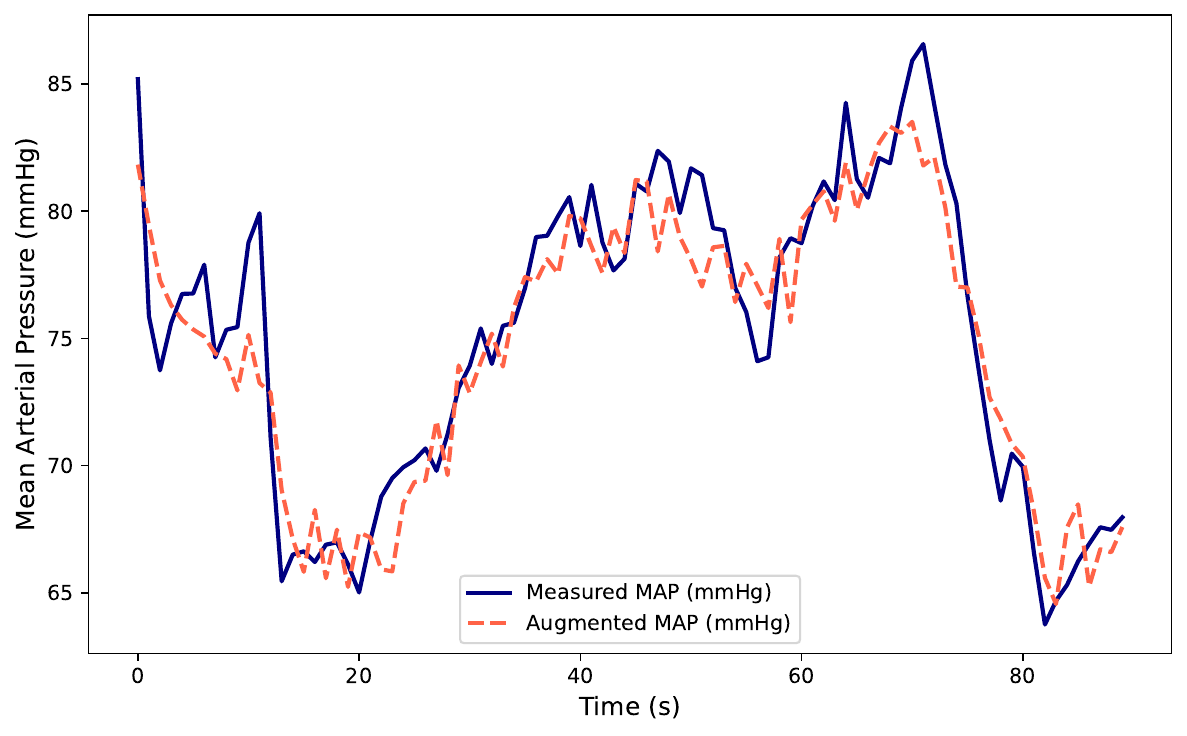}
    \caption{Sample 3 (10s)}
  \end{subfigure}

  \caption{Examples of augmented MAP series of MTRDA under different sampling frequencies.}
  \label{fig:sampling_comparison}
\end{figure}

\section{Hyperparameter Sensitivity}

\begin{table}[htbp]
  \centering
  \caption{Performance of the model under different historical and predicted lengths.}
  \label{tab:difflen}
  \small
  \renewcommand{\arraystretch}{1.2}
  \resizebox{\linewidth}{!}{ 
  \begin{tabular}{ccccccccccccc}
    \toprule
    \multirow{2}{*}{$t$} & \multicolumn{4}{c}{$l=30$} & \multicolumn{4}{c}{$l=60$} & \multicolumn{4}{c}{$l=90$} \\
    \cmidrule(lr){2-5} \cmidrule(lr){6-9} \cmidrule(lr){10-13}
    & MSE$_\mathrm{IOH}$ & MAE$_\mathrm{IOH}$ & Recall & AUC & MSE$_\mathrm{IOH}$ & MAE$_\mathrm{IOH}$ & Recall & AUC & MSE$_\mathrm{IOH}$ & MAE$_\mathrm{IOH}$ & Recall & AUC \\
    \midrule
    30 & $120.85 \pm 11.18$ & $8.98 \pm 0.67$ & 0.482 & 0.6852 & $101.09 \pm 9.34$ & $8.28 \pm 0.64$ & 0.5743 & 0.7288 & $43.29 \pm 8.64$ & $4.65 \pm 0.20$ & 0.7205 & 0.7729 \\
    60 & $72.73 \pm 13.06$ & $6.75 \pm 0.90$ & 0.8191 & 0.7596 & $65.70 \pm 12.91$ & $6.03 \pm 0.82$ & 0.7822 & 0.7444 & $82.07 \pm 11.10$ & $6.88 \pm 0.77$ & 0.8012 & 0.7693 \\
    90 & $56.93 \pm 5.44$ & $5.88 \pm 0.48$ & 0.8368 & 0.7184 & $85.45 \pm 2.79$ & $7.45 \pm 0.45$ & 0.7619 & 0.7318 & $91.16 \pm 15.48$ & $7.25 \pm 0.69$ & 0.7976 & 0.7642 \\
    \bottomrule
  \end{tabular}
  }
\end{table}

To assess the sensitivity of the model to varying historical and predicted horizons, we evaluate performance across different combinations of historical length \( l \) and predicted length \( t \). As shown in Table~\ref{tab:difflen}, increasing the historical length generally improves performance across all metrics. The setting with \( l = 90 \) and \( t = 30 \) achieves the best overall results, with the lowest prediction error, indicating that a longer temporal context enhances short-term IOH risk forecasting. In contrast, extending the predicted length leads to a moderate decline in accuracy, reflecting the increased difficulty of long-range forecasting in clinical settings.

\begin{figure}[h]
    \centering
    \vspace{-0.1in}
    \includegraphics[width=\linewidth]{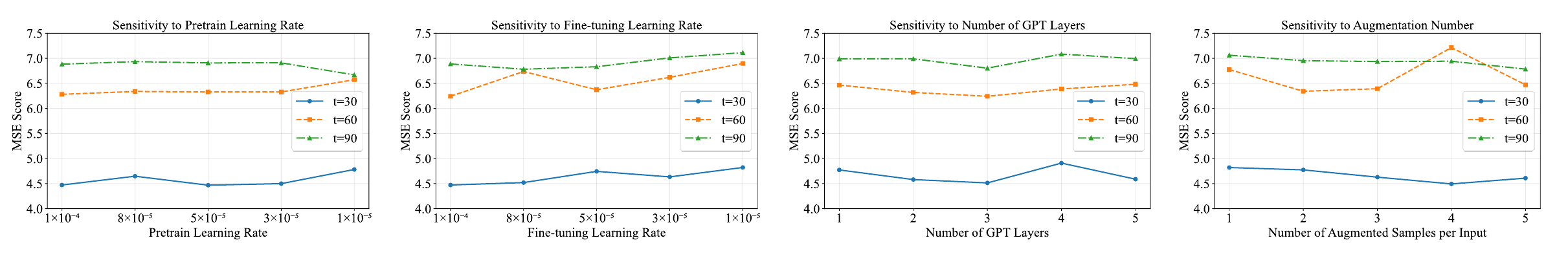}
    \caption{
        Parameter sensitivity analysis on the Clinic IOH dataset.
    }
    \label{fig:Sensitivity}
    \vspace{-0.15in}
\end{figure}

We conduct a sensitivity analysis on the Clinic IOH dataset under 10-second sampling resolution to evaluate the impact of key hyperparameters. As shown in Fig.~\ref{fig:Sensitivity}, the results indicate that model performance is moderately sensitive to the fine-tuning learning rate, while the pretraining learning rate exhibits greater stability. Varying the number of GPT layers shows that moderate depth achieves better generalization, whereas excessive depth may lead to overfitting. Additionally, appropriate levels of data augmented by MTRDA consistently improve performance, though excessive augmentation can introduce distributional noise and degrade accuracy. These findings highlight the importance of balanced model capacity and augmentation strategies for stable performance.

\section{Prompt Design for PCDG}

\begin{wrapfigure}{r}{0.48\textwidth}
    \vspace{-0.1in}  
    \centering
    \includegraphics[width=0.48\textwidth]{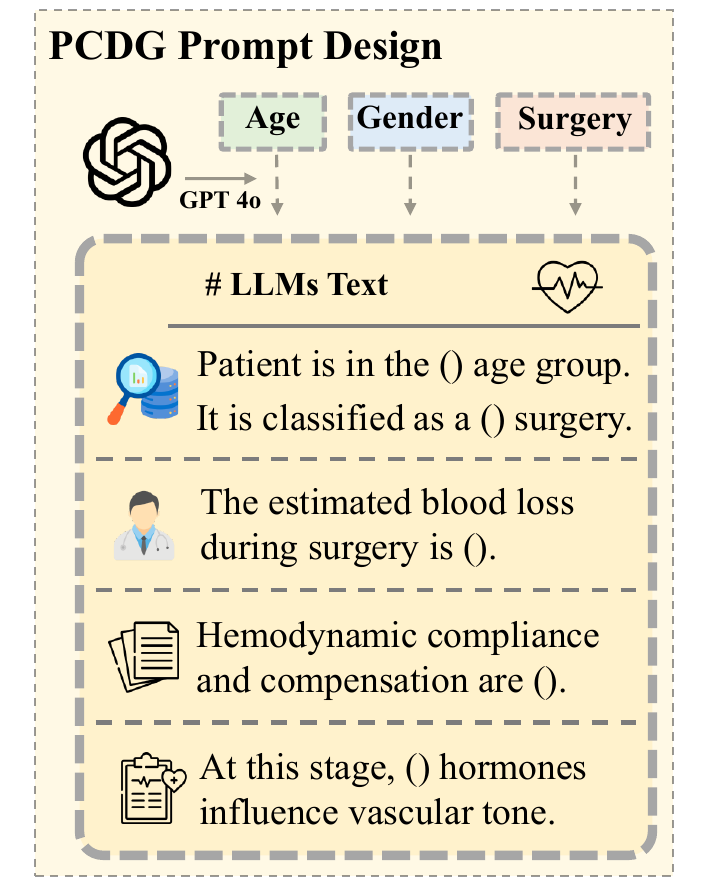}
    \vspace{-0.25in}  
    \caption{Illustration of the PCDG Prompt Design framework.}
    \label{Hyperparameter}
    \vspace{-2.5in}  
\end{wrapfigure}

To generate patient-specific clinical narratives, we design a structured prompt that guides the large language model GPT-4o in producing medically grounded descriptions. This prompt incorporates static patient attributes including age, gender, and surgery type, and aligns with predefined medical templates contextualized by domain knowledge. As shown in Fig.\ref{Hyperparameter}, the generated text serves as a personalized semantic representation used for multimodal fusion in the forecasting pipeline.

\noindent\textbf{Prompt Template:}
\vspace{0.3em}
\begin{tcolorbox}[colback=gray!5!white, colframe=gray!40!black, boxrule=0.3pt, arc=2pt, left=2pt, right=2pt, top=2pt, bottom=2pt]
\small\ttfamily
The age of patient is \{age of patient\}, gender is \{gender of patient\}, and the type of surgery is \{surgery type of patient\}. Please provide the answer directly, separated by commas, without any spaces in between, removing the parentheses when responding. Without any explanations or additional content. The patient belongs to the () age group, whose vascular compliance and cardiovascular compensatory capacity are (). At this time, () hormones act on the blood vessels. This surgery is a () type of surgery, and the blood loss is usually ().
\end{tcolorbox}

This prompt balances consistency with clinical variability by incorporating patient-specific attributes. Grounded in hospital guidelines, clinical heuristics, and literature~\cite{M3, M4}, it is tokenized with an extended vocabulary covering physiological and surgical terms, enabling the model to embed static medical context into prediction.

\section{Broder Impacts}
\label{Broder Impacts}

Although developed for intraoperative hypotension (IOH) prediction, IOHFuseLM is applicable to other clinical tasks involving rare but high-risk physiological events. Similar patterns exist in intraoperative hypoxia detection, where brief desaturation episodes require early identification from noisy SpO$_2$ series, and in intensive care units for sepsis onset prediction, which involves subtle temporal shifts across multiple vital signs. Cardiac arrhythmia monitoring and postoperative respiratory depression detection also share the challenge of aligning transient waveform abnormalities with individualized clinical context.

By combining structured patient descriptions with physiological series, our approach facilitates personalized event recognition in settings where traditional signal-only models may fall short. As such, IOHFuseLM may serve as a general blueprint for multimodal modeling in personalized clinical monitoring systems.

Moreover, as shown in Fig.\ref{fig:Speed_Com}, IOHFuseLM achieves 48ms inference time on an NVIDIA RTX 4090, meeting the responsiveness requirements for real-time physiological monitoring as outlined in ISO 80601-2-77:2017.

\section{Limitations}
\label{Limitations}

While IOHFuseLM shows strong performance on two intraoperative datasets, some limitations remain. The model may be sensitive to differences in data collection protocols across hospitals. Event sparsity and reliance on generated clinical descriptions may also affect its robustness in unfamiliar domains. Future work may explore multi-center pretraining, incorporation of clinical ontologies, or confidence-aware prediction strategies to improve transferability.

\end{document}